\documentclass[conference]{IEEEtran}
\usepackage{times}

\usepackage[numbers]{natbib}
\usepackage{multicol}
\usepackage[bookmarks=true]{hyperref}

\usepackage{amsmath} 
\usepackage{algorithmic}
\usepackage[ruled, linesnumbered]{algorithm2e}
\usepackage{graphicx}
\usepackage{xcolor}
\usepackage[inkscapelatex=false]{svg}
\usepackage{multirow}

\newcommand{\C}{\mathcal{C}}
\newcommand{\D}{\mathcal{D}}

\newcommand{\cfree}{\ensuremath{\mathcal{C}_{free}}}

\DeclareMathOperator*{\argmin}{argmin}

\begin{document}
\title{\LARGE \bf
Path Database Guidance for Motion Planning
}

\author{Amnon Attali$^{1}$, Praval Telagi$^{1}$, Marco Morales$^{1,2}$, and Nancy M. Amato$^{1}$}

\maketitle
\footnotetext[1]{Siebel School of Computing and Data Science at University of Illinois Urbana-Champaign, 201 N Goodwin Ave, Urbana, IL 61801 {\tt\small \{aattali2, ptelag2, moralesa, namato\}@illinois.edu}}
\footnotetext[2]{ ITAM,  México City, México. {\tt\small\{marco.morales\}@itam.mx.}}

\begin{abstract}
One approach to using prior experience in robot motion planning is to store solutions to previously seen problems in a database of paths. Methods that use such databases are characterized by how they query for a path and how they use queries given a new problem. In this work we present a new method, Path Database Guidance (PDG), which innovates on existing work in two ways. First, we use the database to compute a heuristic for determining which nodes of a search tree to expand, in contrast to prior work which generally pastes the (possibly transformed) queried path or uses it to bias a sampling distribution. We demonstrate that this makes our method more easily composable with other search methods by dynamically interleaving exploration according to a baseline algorithm with exploitation of the database guidance. Second, in contrast to other methods that treat the database as a single fixed prior, our database (and thus our queried heuristic) updates as we search the implicitly defined robot configuration space. We experimentally demonstrate the effectiveness of PDG in a variety of explicitly defined environment distributions in simulation. 

\end{abstract}

\IEEEpeerreviewmaketitle

\section{Introduction}

Sampling-based motion planning (SBMP) algorithms use heuristics to guide search in the continuous robot configuration space (CSpace). This guidance commonly comes from ideas of expansiveness~\cite{lavalle1998rapidly, hlm-ppecs-97}, 
laziness~\cite{BohlinK00, Hauser15}, or workspace guidance~\cite{holleman2000framework, DennySBA16}. 

Such methods share some important features, namely that this guidance is quick to compute and updates as the implicitly defined CSpace is explored through collision checking. Another shared property is that they plan from scratch, making very few if no assumptions about the structure of the CSpace. Such generality can be desirable for its simplicity but is also limited in that it doesn't take advantage of the more realistic scenario in which some information is already known about the search space. A simple example is that if certain obstacles in the environment are known to be static (e.g., the table to which a manipulator is attached), collision with these obstacles can be precomputed and used by the planner~\cite{jaillet2004prm, lien2009planning}.

In this work we consider the setting in which offline we have access to samples from a distribution of environments, and must then plan online for a new environment and task sampled from this same distribution. Within reason, we are not concerned with runtime offline, but only how quickly (or with how little collision checking) we can solve this new task. One approach for encoding prior knowledge is to save solutions to previously seen problems as a database of paths (e.g.,~\cite{berenson2012robot}). Algorithms which follow this approach are differentiated by how they answer the following key questions: which paths are kept in the database, how is the database queried, and how is the queried path used (i.e., ``repaired'') in the current CSpace. One additional universal feature is the use of some baseline planning from scratch (PFS) module such as RRT~\cite{lavalle1998rapidly} to provide guarantees in case the database proves insufficient.

Our main contribution is a path-database motion planner, Path Database Guidance (PDG). In contrast to prior work, which queries the database \textit{once} with \textit{task}-based information, PDG does so continuously using information gathered through collision checking. In other words, our database is iteratively filtered throughout the course of the search. Moreover, rather than directly pasting (e.g.,~\cite{berenson2012robot}) queried paths into the current CSpace or using them to learn sampling distributions (e.g.,~\cite{chamzas2022learning}), we use the length of the queried path as an estimate of cost-to-go, i.e., as a heuristic to guide search tree growth. This means that rather than doing PFS in parallel to database guidance or sequentially as two distinct phases, we dynamically alternate between using the database or the PFS module to select which node to expand in an explore-exploit loop.

Our perspective is that the specific set of paths in the database is a hyper-parameter, and thus we don't provide a rule for determining which paths are kept in the database but rather propose that this hyper-parameter be determined through cross-validation on the offline dataset, as is standard in any learning-based approach. We evaluate PDG against both PFS baselines and standard existing database planners, whose parameters were all also tuned on the training dataset. The four algorithms are compared on three CSpace distributions for very different robots, demonstrating that PDG is fastest, most consistent, and uses $2-20$ times fewer collision checks.

\section{Background}

A robot motion planning (MP) problem consists of the tuple $M=(r,E,s,t)$, for robot $r$, environment $E$, and task $(s,t)$. Together, $(r,E)$ describe the configuration space $\C$, a continuous space consisting of all possible configurations of $r$ in $E$. This space is implicitly partitioned into two parts, $\C_{free}$ and $\C_{obs}$. $\C_{free}$ is the set of all valid configurations, and $\C_{obs} = \C \setminus \C_{free}$. A motion planner produces a valid path $p_{s,t}$ from starting configuration $s$ to goal configuration $t$, if one exists. Single query planners generally build a search tree $T$ in $\C_{free}$ by iteratively selecting a node and expanding it, using the primitive operation of collision checking (CC), in which a black box query provides the validity of a single configuration. Edges between nodes of the tree are validated by CC of sufficiently dense discrete configurations along them.

\begin{figure*}
    \centering
    \includegraphics[width=0.3\linewidth]{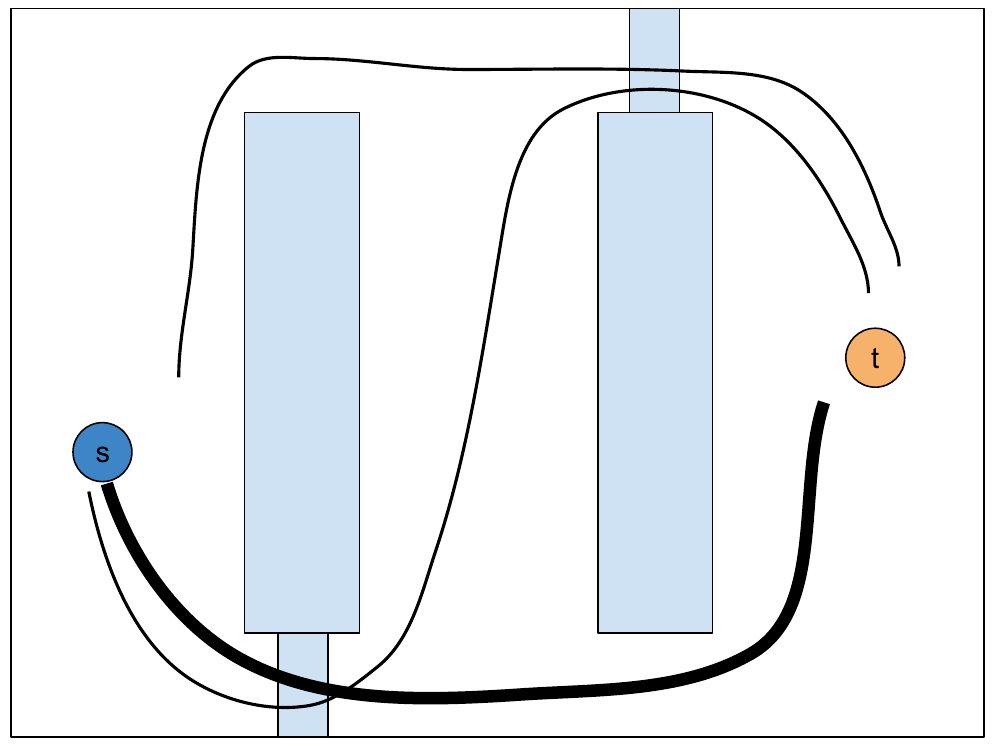}
    \hfill
    \includegraphics[width=0.3\linewidth]{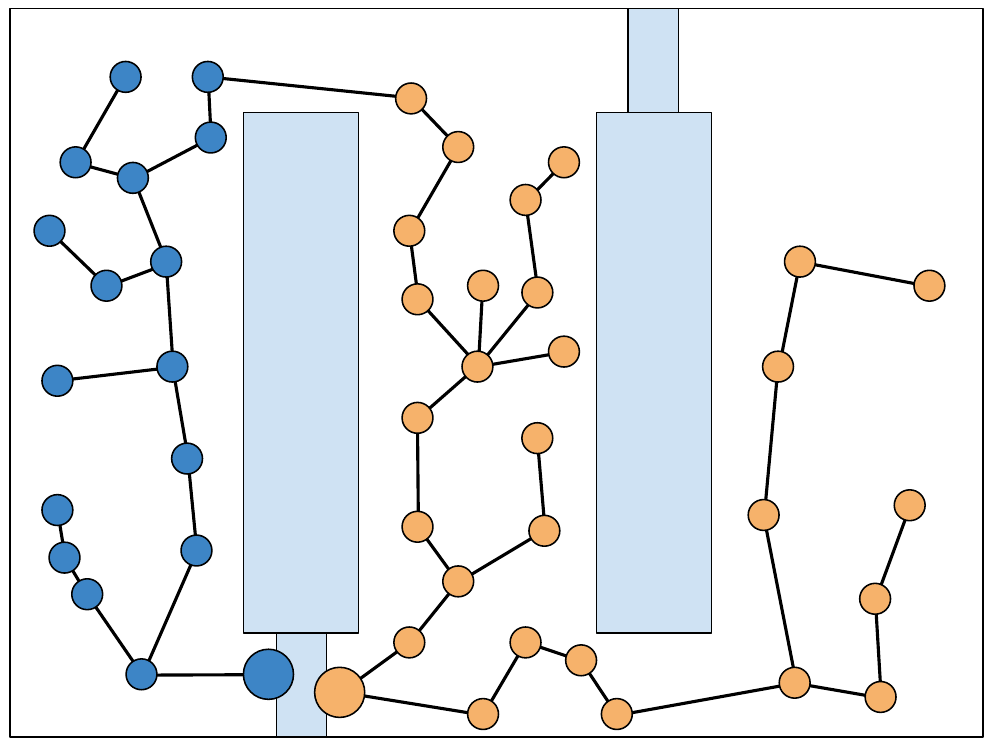}
    \hfill
    \includegraphics[width=0.3\linewidth]{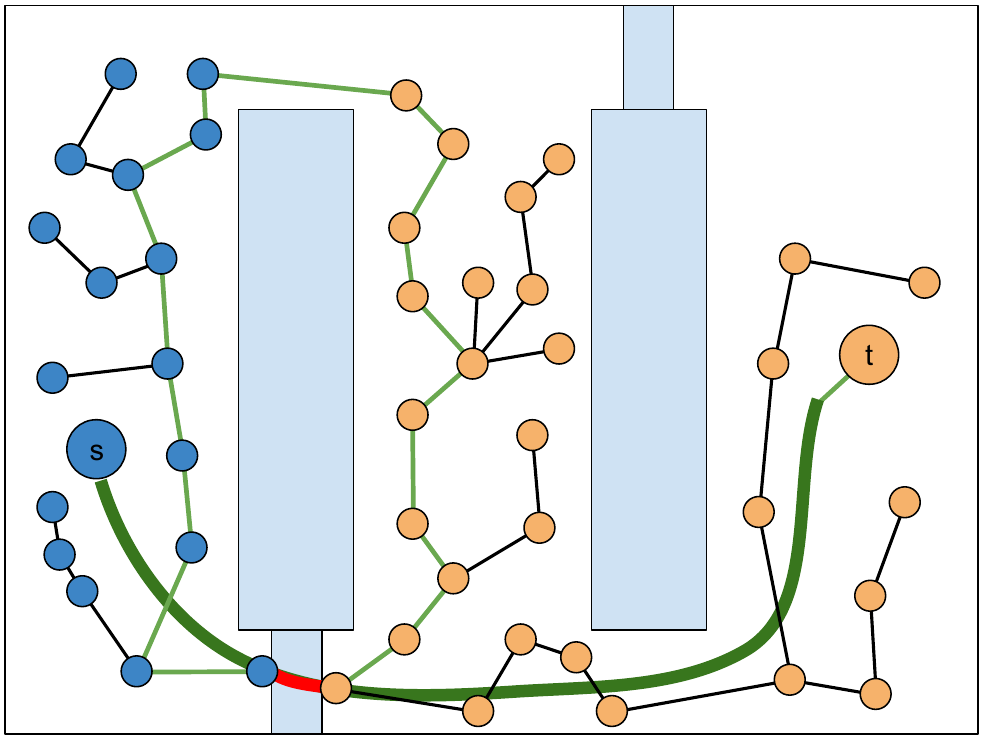}
    \caption{A simple example where Lightning performs poorly despite containing useful paths in its database. In this scenario where no single path solves the $(s,t)$ planning problem on its own, Lightning chooses the path (\textbf{bolded}) with fewest collisions, throwing out all computation relating to the discarded paths (left). Since the entire difficulty of such a problem is encapsulated in navigating this narrow passage, repairing the broken segment with Bidirectional-RRT is approximately as difficult as solving the original problem (center). Finally the original path is combined with the repaired portion to solve the MP problem such that ultimately the database and PFS were used \textit{sequentially} (right).}
    \label{fig:lightning_bad}
\end{figure*}

There are many reasonable ways to evaluate a planner: 
\begin{itemize}
    \item Runtime: a common and practical metric, but which can often depend on implementation details
    \item Path length: we do not focus on this metric since in practice this quantity is often optimized in post-processing via smoothing operations 
    \item Number of CC configurations: CC is usually the most expensive component of motion planning, and also conceptually represents the amount of explicit information needed about the implicit $\C$ to find a path
    \item Number of CC batch calls: many collision checkers are parallelized, meaning they scale sub-linearly with the number of configurations queried in batch but also have some overhead, such as moving data onto a GPU 
\end{itemize}

Like \cite{posterselfcite}, in this work we consider the guided motion planning (GMP) problem, in which we have access to an offline dataset $D = \{M_i \sim \mathcal{D} \}_{i=1}^{n}$ of $n$ MP problems sampled from some unknown distribution $\mathcal{D}$. Our objective is then to produce a guiding space (i.e., data structure) that helps minimize some or all of the above metrics on a new sample $M\sim\mathcal{D}$. In our case we produce a path-database $D_P$ as our guiding space and assume $\mathcal{D}$ to be stochastic over environments and tasks but contain a single fixed robot. We also assume that all environments, while varying in obstacles and obstacle locations, share a single workspace boundary. We leave extensions to more general distributions for future work but note that these assumptions still cover a wide range of real world applications such as fixed-base manipulators and warehouse environments. In our experiments, tasks are either sampled uniformly from $\C_{free}$ or uniformly from a fixed set of configurations (namely solutions to a set of inverse kinematics problems). 

\section{Related Work}

We start with an in depth description of Lightning~\cite{berenson2012robot}, the canonical path-database motion planner. Then we discuss follow-up work and similar path-database planners, such as methods for learning sampling distributions and adapting path-databases to trajectory optimization. 

\subsection{Lightning}

\subsubsection{Path-Database} Given a path-database $D_P$ and a path-based distance metric, i.e., Dynamic Time Warping (DTW), Lightning adds a new path to the database if it is at least some constant threshold distance from any existing path or if a PFS module finds that path faster than Lightning with the current database. In fact the algorithm itself calls for always running PFS in parallel to the database query described below.

\subsubsection{Query/Retrieve} Given a task $(s,t)$ in CSpace $\C$, Lightning queries $D_P$ for the $k$ paths that have most similar starts and goals, in other words it computes the $k$ nearest neighbors in the database according the distance metric $d((s,t), (s',t')) = ||s-s'|| + ||t-t'||$ for some appropriate norm (e.g., euclidean). Those paths are then validated in $\C$, and the path $p$ with fewest collisions (i.e., constraint violations) is returned as the database query.

\subsubsection{Repair} Given a path $p$ that has been retrieved (and validated) from $D_P$, Lightning now repairs invalid segments of the path by running PFS between valid segments.   

\subsubsection{PFS} For PFS they propose BiDirectional-RRT~\cite{kuffner2000rrt}

\subsubsection{(Our) Criticism} Lightning \textit{queries the database once} for a set of $k$ paths using \textit{only task-based information}. The rest of the database is unused, despite potentially containing useful guidance for local sub-problems that must be solved between the start and goal. In other words, the difficulty of motion planning is known to depend heavily on the existence and narrowness of narrow passages, but Lightning provides no guarantees that the queried paths successfully navigate through the narrow passages of the current CSpace, and can thus cause the algorithm to reduce to running BiRRT between configurations on either side of a narrow passage. While CSpace information is subsequently incorporated into the query by fully validating the $k$ paths, this not only scales poorly with $k$ but also cannot make up for a poor initial filtering. See Figure~\ref{fig:lightning_bad} for a visual example of a natural situation in which Lightning performs poorly. 

\begin{figure*}[t]
    \centering
    \includegraphics[height=0.18\linewidth, width=0.24\linewidth]{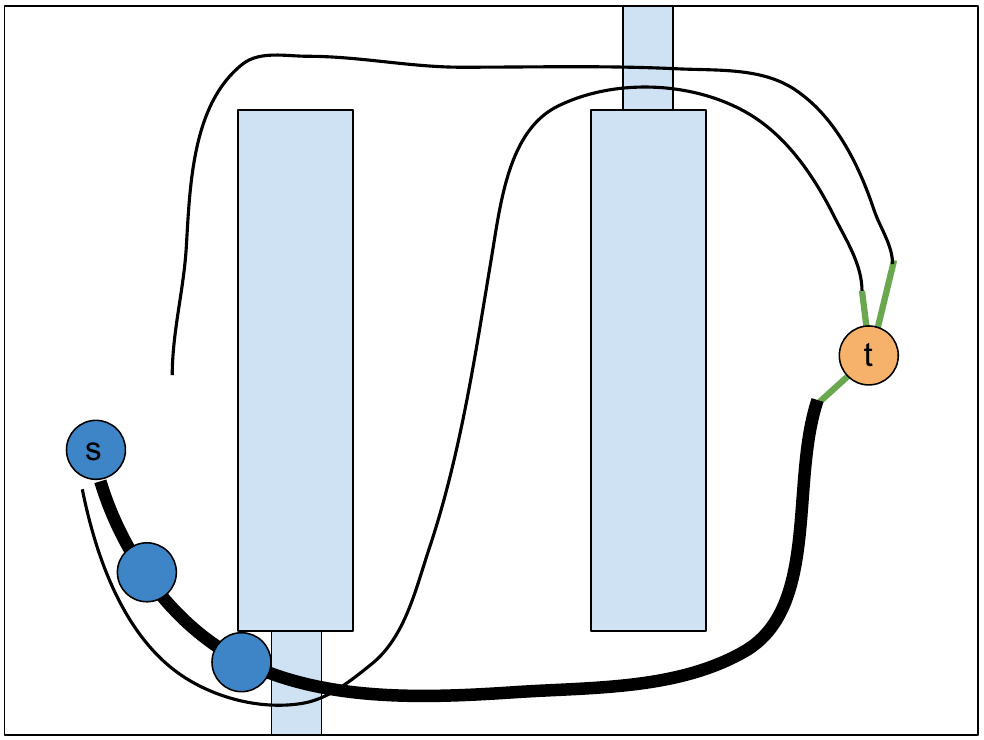}
    \hfill
    \includegraphics[height=0.18\linewidth, width=0.24\linewidth]{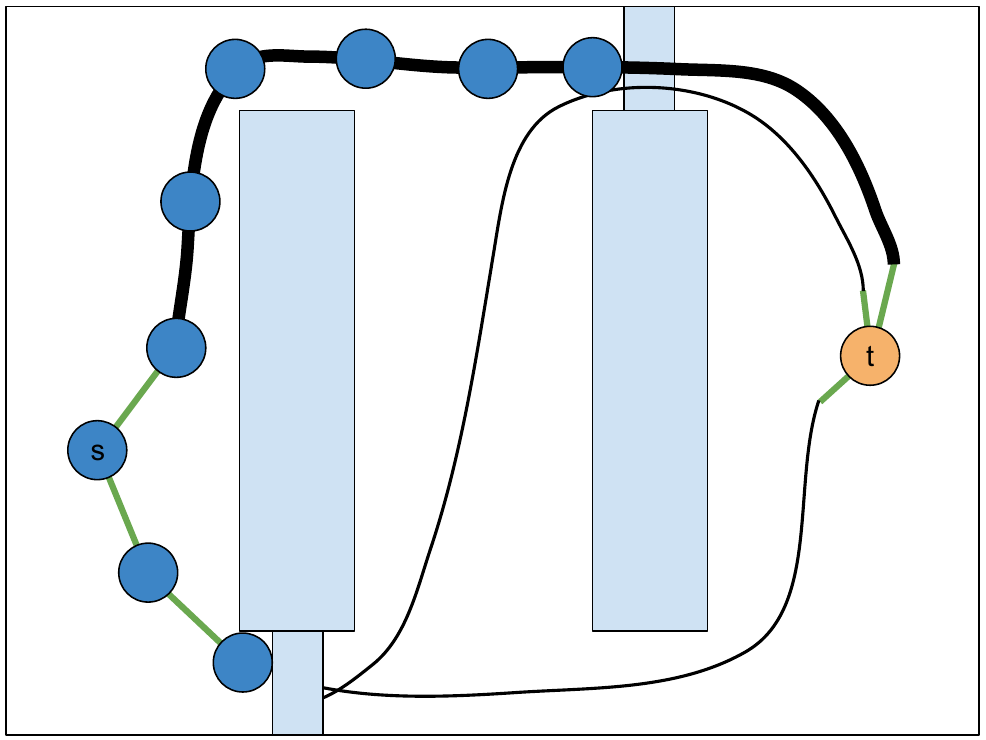}
    \hfill
    \includegraphics[height=0.18\linewidth, width=0.24\linewidth]{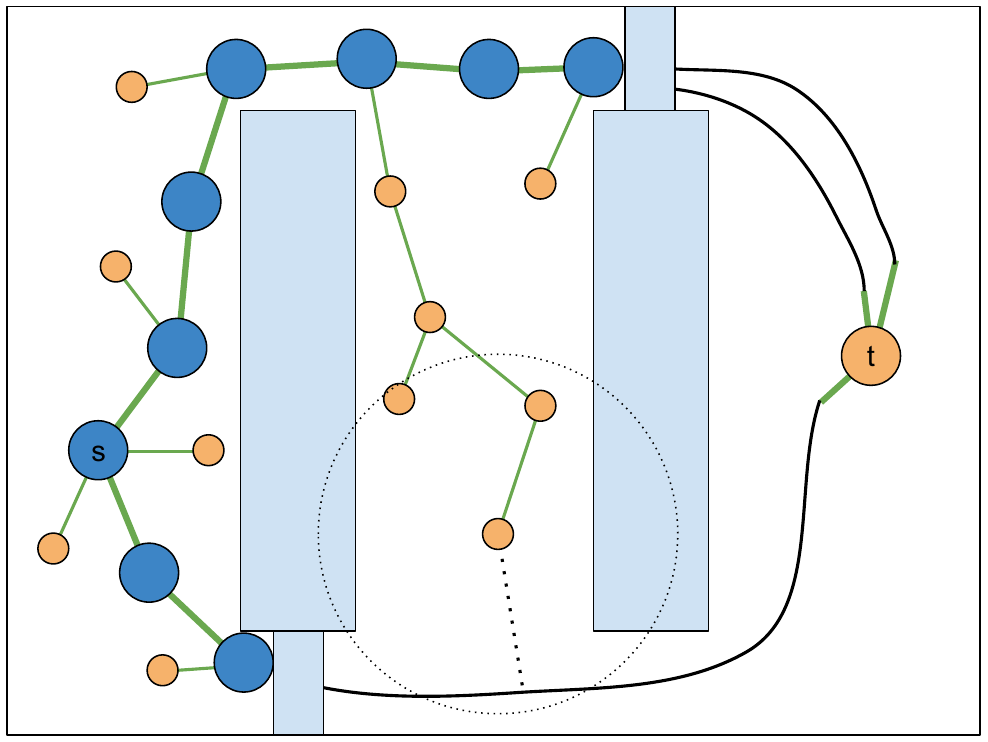}
    \hfill
    \includegraphics[height=0.18\linewidth, width=0.24\linewidth]{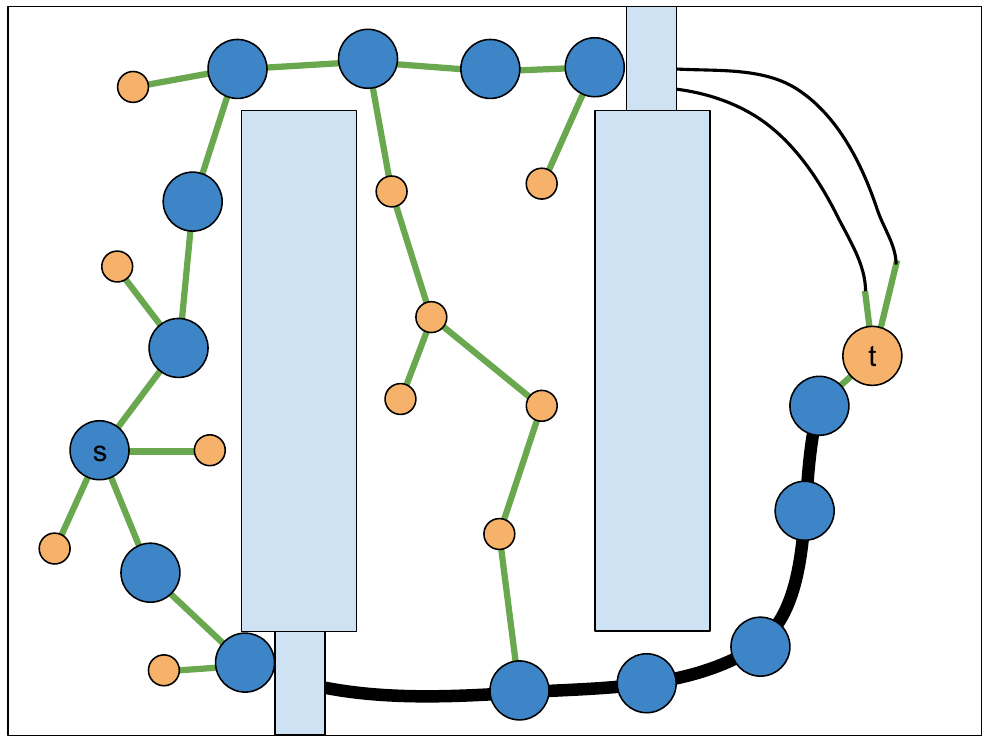}
    \caption{Example run of PDG on the same environment and paths as Figure~\ref{fig:lightning_bad}. Validated edges are marked green. First we filter the paths to those that pass near the goal $t$, and validate the connection to $t$. Next we attach from the search tree (the start state $s$) onto a nearby path and follow it until we collide with an obstacle (left). The part of the path before the invalid portion is deleted and we attach onto the next available path (center left). If there is no nearby path to attach to we select which node from the tree to expand using RRT until we can again latch onto a path (center right). This process continues until we reach the goal (right).}
    \label{fig:pdg_example}
\end{figure*}

\subsection{Experience-Based Planners}

Lightning has inspired a host of planners that follow a similar framework. \cite{stolle2007transfer} uses a hand designed distance metric to query a library of footsteps (for a quadruped) and then use a local optimizer to connect disconnected segments. Thunder~\cite{coleman2015experience} extended Lightning by storing experiences in a sparse roadmap spanner rather than as individual paths, thus generalizing the concept of not storing paths that are too similar, yet they do not address the concept of roadmap sparsity \textit{across environments}. ERT~\cite{pairet2021path} focuses on using a single path from a database by deforming local sections of it to yield an algorithm that is reminiscent of path tracking for some horizon with model predictive control. Similarly, \cite{elimelech2022efficient} adapts a small database of abstract plans through geometric deformations to the current problem instance. Like us, both E-Graphs~\cite{phillips2012graphs} and MHA*-L~\cite{aine2015learning} use the length of a path as a heuristic for guiding search, but both methods are designed for efficiently planning for a new task in a \textit{similar} environment, making them more closely related to graph sparsification methods such as SPARS~\cite{dobson2014sparse} than to our problem setting. 

One line of related work produces a learned sampling distribution for extending a search tree. In other words, rather than sampling $\C$ uniformly (e.g., RRT) samples are drawn from this learned distribution. SPARK, FIRE, and FLAME~\cite{chamzas2021learning, chamzas2022learning} use handcrafted or learned similarity metrics over local workspace occupancy to query a database for a global sampler. Like the Repetition Roadmap~\cite{lehner2018repetition}, this sampler is defined as a Gaussian-Mixture Model where each individual gaussian is centered at some critical configuration. Some deep learning approaches follow this same paradigm of learning a sampling distribution over CSpace given some global representation of the environment such as a top down image~\cite{qureshi2019motion, ichter2019robot}. Note that such distributions, while they do take into account some information related to the current CSpace (e.g., through an image of the environment), are nevertheless \textit{fixed priors that do not update as search progresses.}

Trajectory optimization approaches often describe problem distributions in terms of ``p-parameters'', such that the algorithm focus is on database \textit{coverage} of this parameter domain rather than query and adaptation~\cite{hauser2016learning, ishida2024coverlib}. Other approaches treat queries from the database as warm-starts for the optimization, where much of the focus is on designing appropriate features so that a single nearest neighbor query yields a good seed~\cite{jetchev2013fast, lembono2020memory}. Again, notice that such methods \textit{query the database once} and subsequently use a baseline adaptation algorithm to fix unsatisfied constraints.

\begin{figure*}[ht]
    \centering
    \includegraphics[height=0.2\linewidth,width=0.3\linewidth]{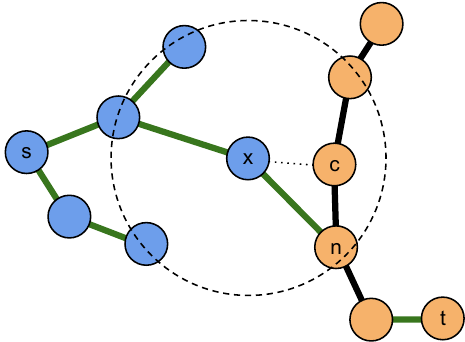}
    \includegraphics[height=0.2\linewidth,width=0.3\linewidth]{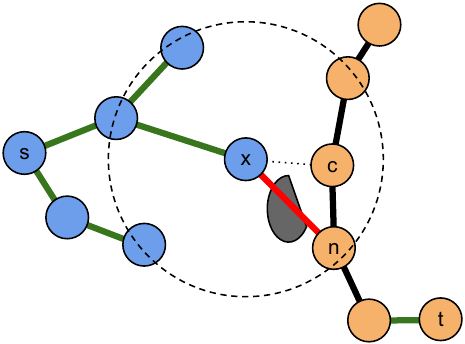}
    \includegraphics[height=0.2\linewidth,width=0.3\linewidth]{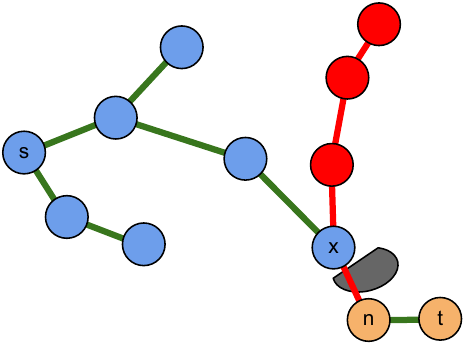}
    \caption{For computing the value of state $x$ we find the closest state $c$ on every path, if $c$ is within $\delta$ of $x$ then we validate the edge to the next state after $c$ in the path, $n$. If that connection is valid (left) we say the estimated cost-to-go value of $x$ is the distance from $x$ to $n$ plus the distance from $n$ to the end of the path. If the edge is invalid (middle) our estimate for $x$ (from this path) is infinite. If $x$ is itself on the path (right) but the edge to $n$ is invalid then we delete the part of the path up until the invalidated portion.}
    \label{fig:alg_ops}
\end{figure*}

\section{Method}

Our algorithm PDG (see Algorithm~\ref{pdgs_alg} for pseudo-code and Figure~\ref{fig:pdg_example} for an example run on a simple environment) follows the guiding space framework~\cite{posterselfcite} in which a search tree is iteratively expanded in CSpace by querying a guiding space for the heuristic value of each node. At each iteration, the node with minimum heuristic value (see Section~\ref{sec:query}) according to the path-database is selected for expansion. As we compute this heuristic value we do collision checking, which allows us to update (see Section~\ref{sec:update}) the path database by removing invalid path segments. If no node can attach onto a path in the database we default to expanding the tree according to a baseline exploration method, namely RRT (Alg~\ref{pdgs_alg} lines 5-6). 

Conceptually what we do is similar to LazyPRM~\cite{BohlinK00}, except instead of validating edges of a graph we validate edges of paths in a database, and like LazyPRM we prioritize paths which lead the robot to the goal the fastest. Where LazyPRM initializes by building an unvalidated graph, we initialize by filtering our database to potentially useful paths.

\subsection{Creating the path-database}

\begin{algorithm}[t]
    \SetAlgoLined
    \SetKwInOut{Input}{Input}
    \SetKwInOut{Init}{Init}
    \Input{GMP $(M = (r, E, s, t),$ Path-database $D_P)$, $\delta$}
    Initialize search tree $T=\{s\}$\\
    Filter $D_P$ to paths near goal: $D^t_P = \{(p_0, \dots, p_k, t) : p\in D_P, p_k \in B_\delta(t), (p_k, t) \in \cfree \}$\\
    \While{$t \not\in T$}{
        $\displaystyle v_{select} = \argmin_{v\in T} V_D^\delta(v)$ \tcp{QUERY+UPDATE}
        \eIf {$V_D^\delta(v_{select}) = \infty$}{
        Expand tree according to RRT \tcp{EXPLORE}}{
        $\displaystyle p = \argmin_{p \in B_\delta (v_{select})} V_p(v_{select})$ \tcp{EXPLOIT}
        Add $x_c^p$ to $T$ as child of $v_{select}$ }
    }
    \caption{Path Database Guided Search}
    \label{pdgs_alg}
\end{algorithm}

\subsubsection{Offline Training}
Given a GMP in the form of a dataset $D = \{(r, E_i, s_i, t_i)) \sim \mathcal{D} \}_{i=0}^n$ of MP problems we create a path-database $D_P$ using an asymptotically optimal motion planner (PRM*~\cite{karaman2011sampling}) to plan paths between $s_i$ and $t_i$. We then smooth the paths by skipping vertices if the edge from the previous to the next path vertices is valid. The problem definition above implies that the max size of our path-database depends on the size of the GMP database, but in practice a dataset usually consists of a set of environments and it is up to us (the algorithm) to determine how many tasks to sample. In other words, $D = \{(r, E_i) \sim \D_E\}_{i=0}^n$ for some environment distribution, and we have access to unlimited task samples $(s, t) \sim \D_{task}(E_i)$ for some task distribution dependent on the environment. In general, regardless of whether tasks have been fixed or not, we propose to determine which paths to keep in the database by validating subsets of paths on $D$ itself (or some held out set), via classical cross-validation techniques. See Section~\ref{sec:path_db} for a discussion on how we do this in our experiments. 

\subsubsection{Online Initialization}
Given a new MP problem $M = (r, E, s, t) \sim \D$ we initialize a search tree at the start $s$ (Alg~\ref{pdgs_alg} line 1) and filter $D_P$ to the set of paths that pass near the goal $t$ (line 2). Our objective is to create an active subset of paths such that if our search tree reaches the end of any of those paths then we have found the goal. We could do this by validating every straight-line edge from every intermediate state on every path to the goal, but this is obviously prohibitively expensive. Instead, for every path we find the closest state on that path to the goal and only validate that edge. Moreover, to avoid validating long (global) edges which are more expensive, we only validate those that are within $\delta$ of the goal. 

\begin{gather*} 
D^t_P = \{(p_0, p_1, \dots, p_{k-1}, t)\} : \\
p\in D_P\\
d(p_{k-1}, t) \leq d(p_j, t)\:\forall j\neq k-1\\
p_{k-1} \in B_\delta(t)\\
(p_{k-1}, t) \in \cfree
\end{gather*}
where $d(a, b)$ is the straight line (i.e., local planner) distance between states $a$ and $b$, and $B_\delta(t)$ is a $\delta$-radius ball centered at $t$. Notice already a departure from Lightning - whereas Lightning proposes to treat paths as fixed objects with a start and end, we make use of paths that pass near the goal but may end up far from it. While this operation is more expensive than comparing distance to path endpoints we find it inconsequential as part of the whole.

\begin{figure*}[ht]
    \centering
    \includegraphics[height=0.2\linewidth,width=0.3\linewidth]{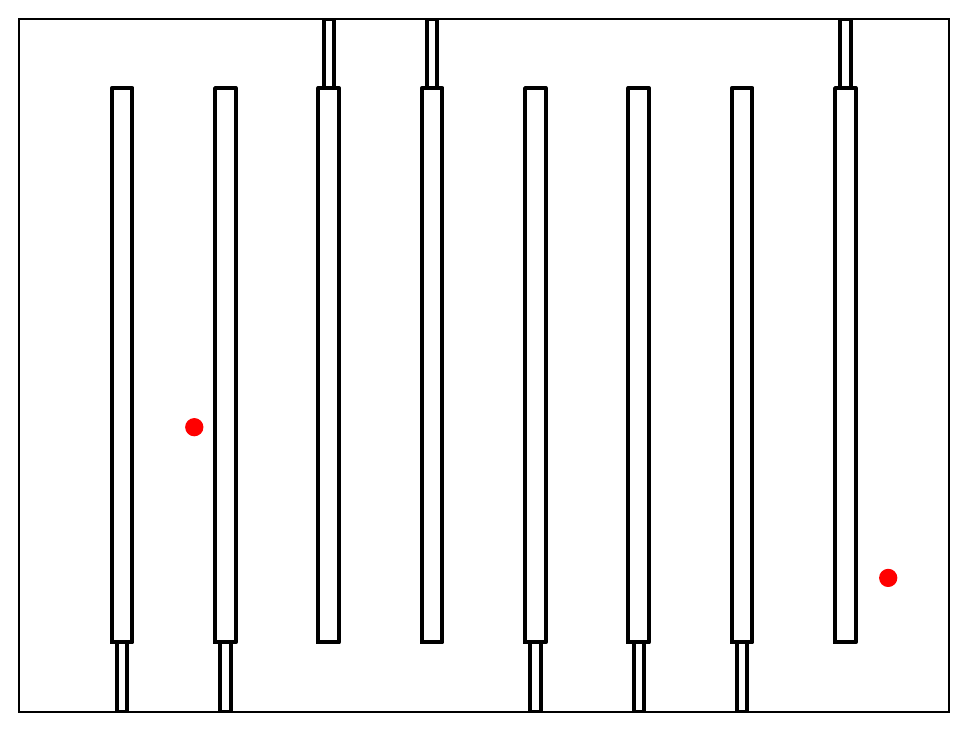}
    \includegraphics[height=0.2\linewidth,width=0.3\linewidth]{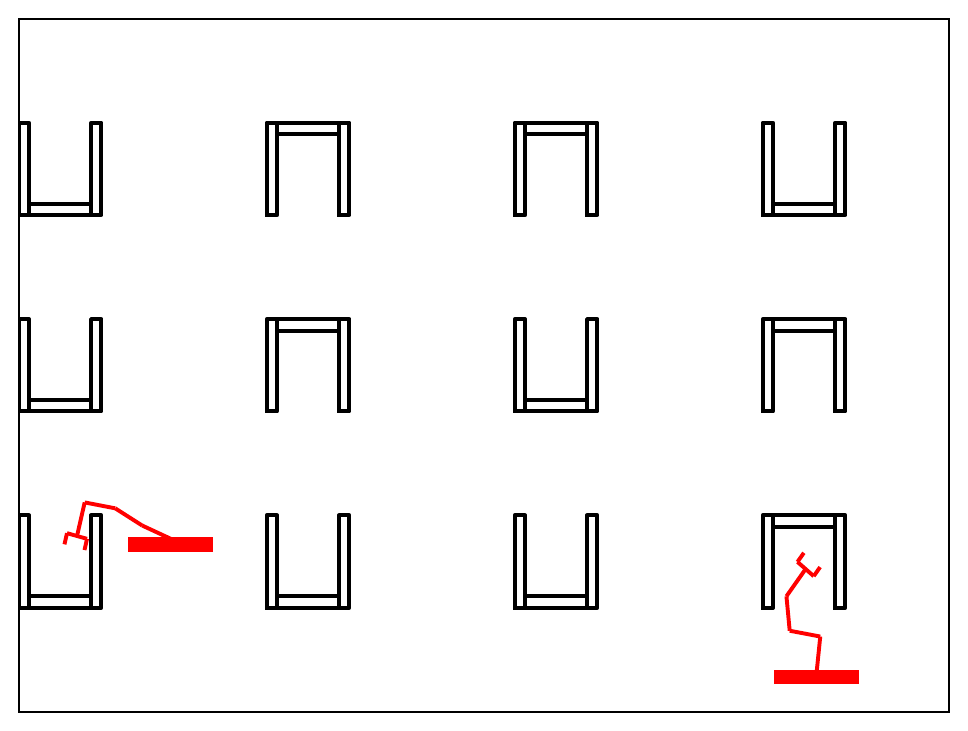}
    \includegraphics[height=0.2\linewidth,width=0.3\linewidth]{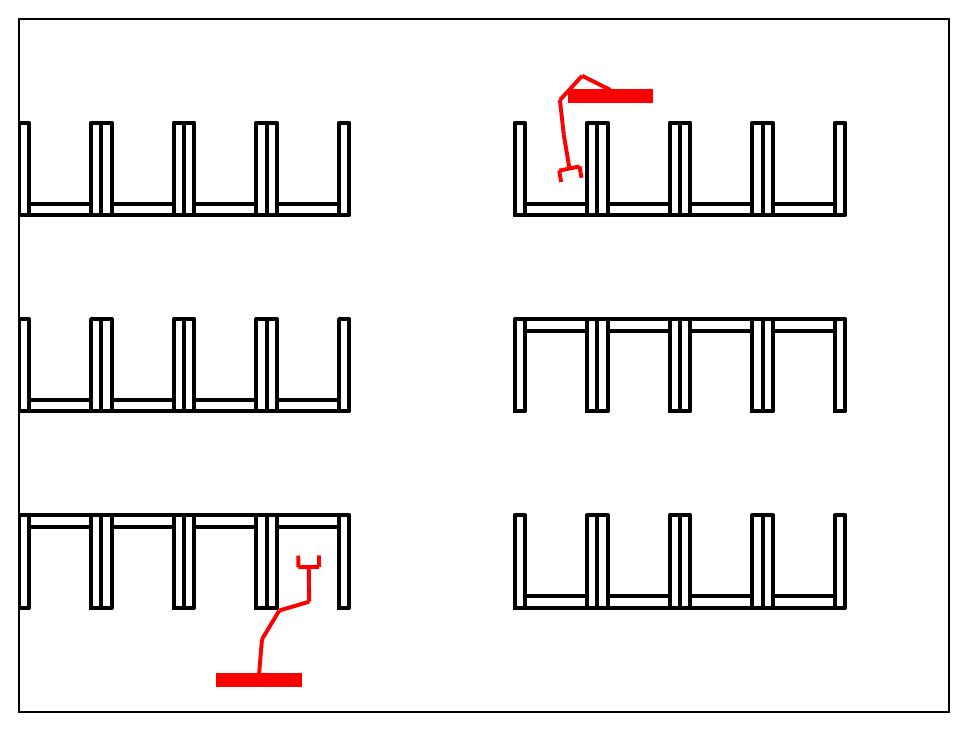}\\
    \includegraphics[height=0.3\linewidth, width=0.4\linewidth]{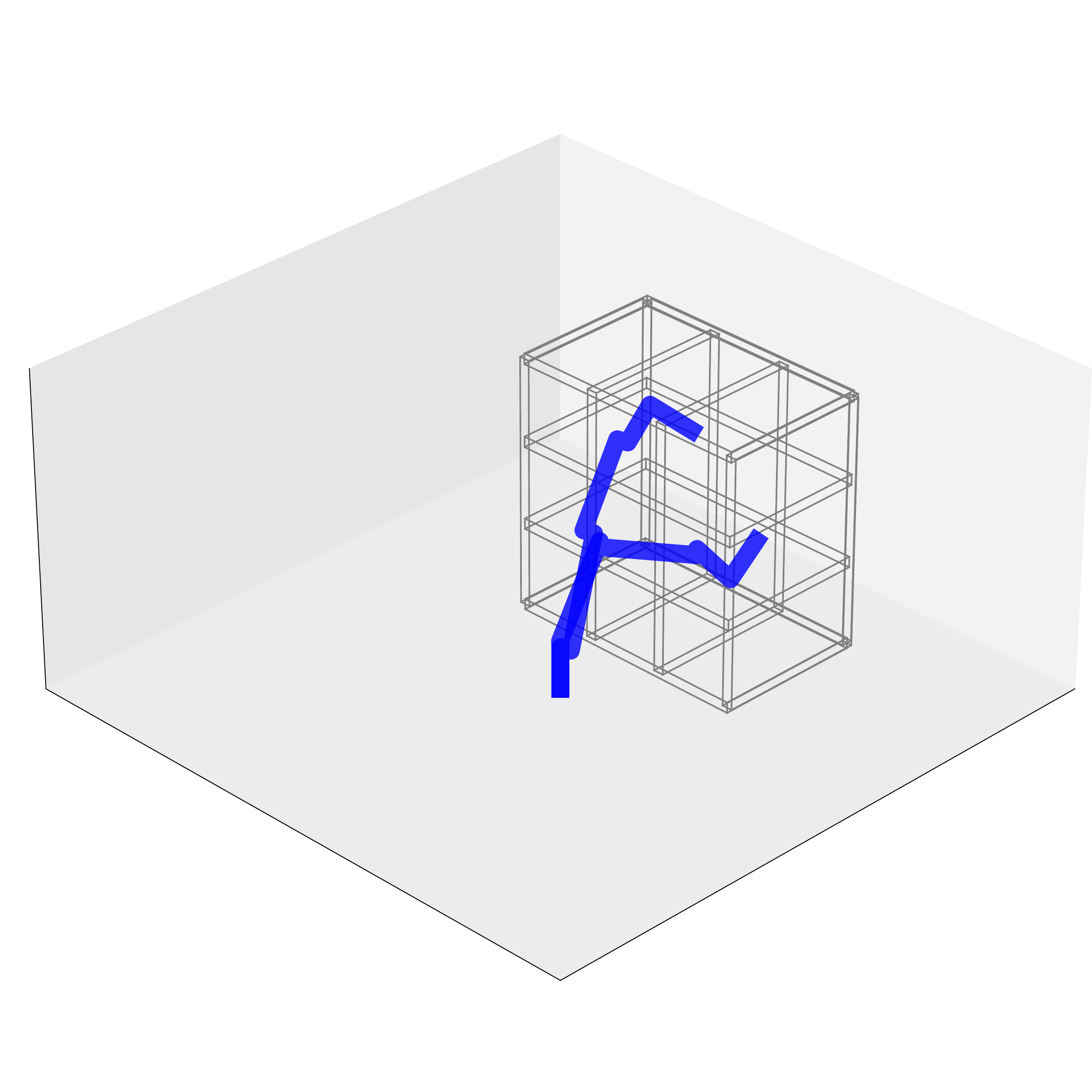}
    \hspace{1cm}
    \includegraphics[height=0.3\linewidth, width=0.4\linewidth]{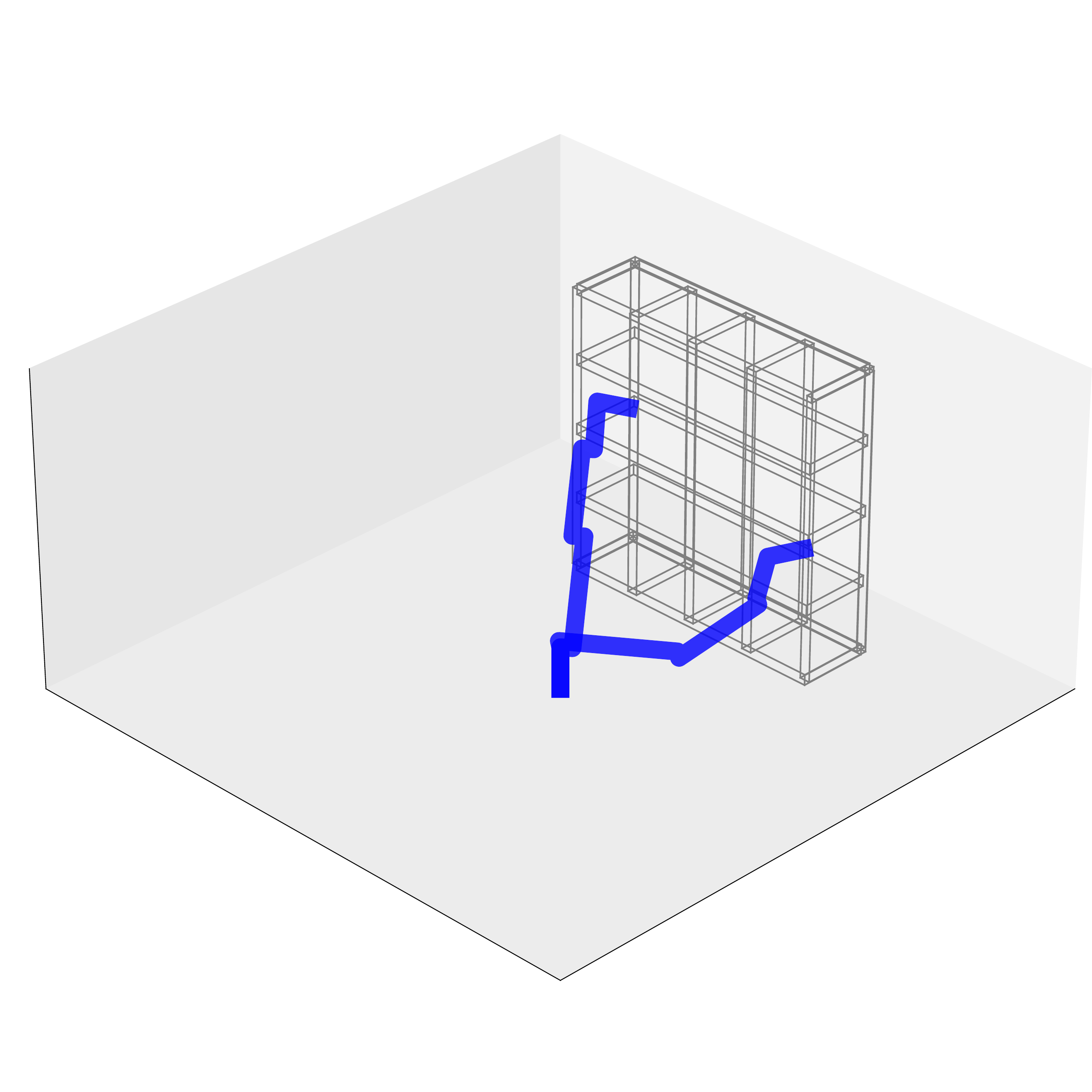}
    \caption{Example samples of environments and tasks from our three distributions, RandomPassage (top left), Cubicles (top/center right), and Shelves (bottom). In RandomPassage each vertical obstacle is blocked at the top or bottom. In Cubicles each group of cubicles faces the same direction but the number and group orientation are sampled at random. In Shelves the number, distance, size, and depth of the shelves varies, with the task always sampled as a configuration whose end effector is in the center of a shelf.}
    \label{fig:env_examples}
\end{figure*}

\subsection{Query (line 4)}
\label{sec:query}

See Figure~\ref{fig:alg_ops} for examples of how our algorithm assigns heuristic values to states by querying the database. 

Given a path $p = (p_0, p_1, \dots, p_k = t)$, we say the path-optimal-value of state $x$ is the minimum distance to get from $x$ to $t$ if one were to first travel from $x$ to $p$ and subsequently move along $p$ to $t$:
$$V^{opt}_p(x) = \min_{p_i \in p}
\begin{cases} 
    d(x, p_i) + c(p_i) & edge(x, p_i) \subseteq \cfree\\
    \infty & otherwise
\end{cases}$$ 
where $c(p_i) = \sum_{j>i} d(p_{j-1}, p_j)$, in other words the distance along the path from $p_i$ to $p_k = t$. Notice that we only validate the edge from $x$ to the path and not the edges of the path states, and that by the triangle inequality this operation is equivalent to finding the largest $i$ such that the edge $(x, p_i)$ is valid.

Computing $V^{opt}_p$ can require $k+1$ edge validations, which is quite expensive, and so we approximate this quantity with the much cheaper to compute path-closest-value:
$$V_p(x) =
\begin{cases} 
    d(x, x_{c}^p) + c_p(x_{c}^p) & edge(x, x_{c}^p) \subseteq \cfree\\
    \infty & otherwise
\end{cases}$$
where $\displaystyle x_c^p = \argmin_{i=1,\dots,k} d(x, p_{i-1})$ is the \textit{next state after} the closest state on $p$ to $x$, i.e., the projection of $x$ onto the path. The reason we use the next state and not the closest ensures that if $x$ is on the path then $x_c^p \neq x$. The two cases where the edge is valid and invalid are visualized in the left and center images of Figure~\ref{fig:alg_ops} respectively.

Now, given $D_P$ we say the path-database-value of state $x$ is: $V_{D}(x) = \min_p V_p(x)$. Unfortunately, with a database of $n$ paths this still requires $n$ (global) edge validations, making $V_D$ expensive to compute. So again, we approximate by only considering paths that pass within $\delta$-radius ball around $x$:
\[V^\delta_D(x) = \min_{p} V_p(x) \text{ such that } p \cap B_\delta(x) \neq \emptyset\]
Notice that while this may still require $n$ edge validations, they are all local with length at most $\delta$. In practice our implementation allows for a different $\delta$ for initial goal filtering and subsequent path-value computation. 

\subsection{Update}
\label{sec:update}

When computing $V_p(x)$, if $x$ is on the path, i.e., $x = p_{i-1}\in p$ then the edge $(x, x_c^p) = (p_{i-1}, p_{i})$ is on the path $p$. If this edge is invalid, i.e., $V_p(x) = \infty$, then we update the path by deleting all states up until $p_{i}$. Let 
\[j = \begin{cases} i & p_{i}\in \cfree\\i+1 & p_{i}\not\in\cfree \end{cases}, p \leftarrow (p_j, p_{j+1}, \dots, p_k)\]

By updating the path in this way we ensure that any future guidance along this path must come from a later part of the path, avoiding this invalidated edge $(p_{i-1}, p_i)$. This path update operation is visualized in Figure~\ref{fig:alg_ops} right.

\subsection{Implementation Details}

\subsubsection{Caching} Any time we compute $V_p(x)$ or $V_D^\delta(x)$ for any $x$ we cache these values (and save the associated path $p$). This means the only times we have to compute these values is for new states added to the tree and for any $v \in T$ whose associated path $p$ is updated. In other words we never collision check an edge twice and operations such as line 8 in Algorithm~\ref{pdgs_alg} are constant time.

\subsubsection{Batched operations} Most modern collision checkers are more efficient at validating $m$ configurations in batch rather than doing $m$ individual checks. Similarly, computing all path states within a $\delta$ radius of a new node $v$ is much faster as one batch computation across all paths than as $n$ individual queries. We take advantage of such parallelism where appropriate. 

\subsubsection{Greedy evaluation} The description of the algorithm implies every new state added to the tree should be collision checked for connections to nearby paths. In practice, if a state is \textit{on} some path then we lazily defer full state-valuation until this path is deleted. We keep a stack of newly explored states and once the attached path is updated we pop from this stack until successfully attaching onto a new path or resorting to exploration.

\subsubsection{Choice of RRT for exploration} While any exploration algorithm would work in line 6 we choose RRT because exploration occurs when no path passes close enough to our tree (with a valid edge), and so a Voronoi bias that expands to unexplored regions is a logical choice for coverage. Nevertheless, other exploration methods are reasonable, for example biasing tree growth towards a randomly chosen path.

\section{Path selection as a parameter setting}
\label{sec:path_db}

\begin{figure}
    \centering
    \includegraphics[width=\linewidth]{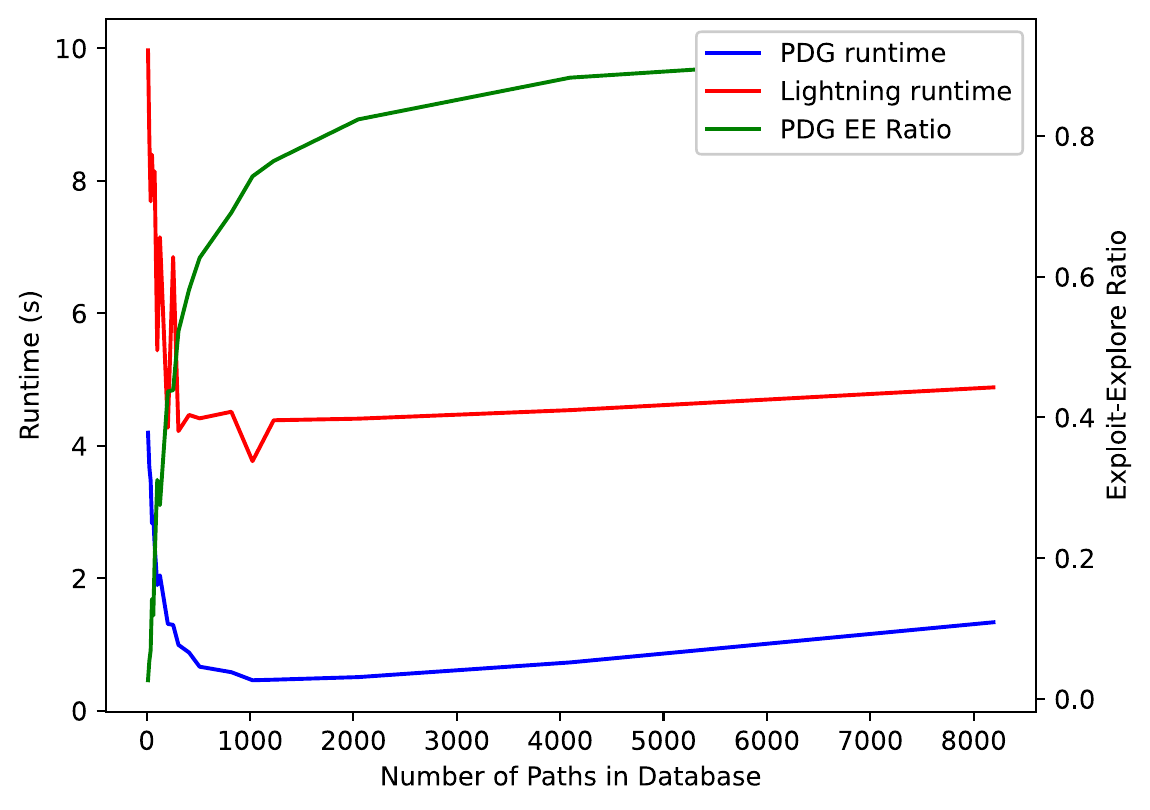}
    \caption{A comparison between Lightning and PDG runtime as a function of the number of paths in the database on the RandomPassage environment~\ref{fig:env_examples}. The Exploit-Explore (EE) ratio is the number of times PDG uses the database to determine which tree node to expand divided by the total number of tree expansions. Both algorithm implementations use (bidirectional) RRT to repair and explore respectively. Notice how Lightning achieves a peak performance around $100$ paths and stays relatively flat as the database grows nearly two orders of magnitude. In contrast, PDG smoothly improves as the database gets larger until around $1000$ paths, and then gets worse as guidance from the database saturates (the EE ratio increases). This matches intuition, that the method pays for more paths than necessary, and that Lightning has a very weak dependence on the database size since most it is unused during each query.}
    \label{fig:performance_size}
\end{figure}

How to select paths for a path database method should be different for each different method. For example, Lightning queries using only task information and so a logical choice is to aim for a path database which, for each start and goal pair has a variety of possible paths between them. This motivates the approach in~\cite{berenson2012robot} to avoid keeping pairs of paths whose DTW is too similar. In their experiments they propose a threshold of $5$. Why $5$? Where does this number come from? In Thunder~\cite{coleman2015experience} they use a stretch factor of $t=1.2$ and compare to a Lightning implementation with DTW threshold of $4$. Where do these numbers come from? In~\cite{aine2015learning}, rather than using a DTW threshold they cluster paths using DTW as a metric. In their experiments they use $4$ clusters of paths that were broken into $5$ segments, and no indication where these numbers come from. SPARK~\cite{chamzas2021learning} contains a table of $7$ parameters for the learned sampling distribution, ``tuned heuristically''.

We propose a simple and general approach that eliminates such parameters by taking advantage of the problem definition itself. In GMP future problems are assumed to be sampled from the same distribution as the training data, and thus parameters can be tuned on the training data to maximize performance on unseen instances. In other words we split our dataset $D = D_{train} \cup D_{validate}$, generate paths exhaustively for all tasks in $D_{train}$, and validate subsets of paths on problems in $D_{validate}$. The subset of paths which yields best validation performance is used for testing. Notice that for a database of size $|D_P|$ there are $2^{|D_P|}$ possible subsets, an infeasible quantity to validate fully.

In our experiments we try both random subsets of paths of different sizes and a DTW based greedy clustering with different thresholds. For each percentage in a range we take $k$ random sub-samples of that percentage. For each clustering threshold in a range, for each value $m$ for number of paths to keep from each cluster, we take $k$ random subsets of $m$ paths from each cluster. We report the exact set of values in Section~\ref{sec:ablation}. In Figure~\ref{fig:performance_size} we show how our validation performance depends on the size of the path database (for random subsets) as compared to Lightning.

\section{Experimental Setup}

\begin{table*}
\centering
\def\arraystretch{1.2}
\begin{tabular}{|c | c | c | c | c | c | c | c || c | c | c | c | c | c|} 
 \hline 
 & & \multicolumn{6}{c|||}{Search Time (s)} & \multicolumn{6}{c|}{Collision Checks (x1000)} \\
 \hline 
 Environment & Algorithm & mean & median & min & max & std & IQR & mean & median & min & max & std & IQR\\
 \hline
 \multirow{7}{*}{RandomPassage} & PRM & 0.65 & 0.58 & 0.30 & 1.47 & 0.36 & 0.54 & 537 & 503 & 321 & 997 & 225 & 357\\
 & BiRSG & 2.68 & 2.14 & 0.25 & 7.07 & 1.75 & 2.66 & 561 & 504 & 74.0 & 1224 & 273 & 399\\
 & PDGRSG & 0.31 & 0.27 & 0.15 & 0.69 & 0.14 & 0.21 & 23.1 & 14.6 &  \textbf{1.37} &  104 & 23.6 &  27.9\\
 & LightningRSG & 1.06 & 0.65 & 0.05 & 10.68 & 1.58 & 1.77 & 173 & 146 & 49.9 &  596 & 124 & 205\\
 & BiRRT & 30.8 & 27.1 & 3.05 & 102 & 22.6 & 31.6 & 310 & 296 & 95.1 & 871 & 157 & 215\\
 & PDGRRT & \textbf{0.15} & \textbf{0.13} & \textbf{0.05} & \textbf{0.37} & \textbf{0.07} & \textbf{0.12} & \textbf{15.8} & \textbf{14.5} & 3.06 & \textbf{44.1} & \textbf{9.53} & \textbf{13.0}\\
 & LightningRRT & 2.70 & 0.78 & 0.03 & 13.78 & 3.86 & 4.75 & 126 & 87.6 & 47.3 & 355 & 82.4 & 142 \\
 \hline
 \multirow{7}{*}{Cubicles} & PRM & 1.07 & 0.82 & 0.31 & 3.48 & 0.71 & 1.24 & 988 & 800 & 247 & 2948 & 636 & 1204\\
 & BiRSG & 1.32 & 1.18 & 0.19 & 3.56 & 0.71 & 0.89 & 369 & 327 & 51.2 &  985 & 192 & 257\\
 & PDGRSG & \textbf{0.89} & \textbf{0.64} & 0.26 & \textbf{3.07} & \textbf{0.66} & 0.99 & 162 & 107 & \textbf{11.3} & 701 & 149 & 194\\
 & LightningRSG & 0.93 & 0.79 & \textbf{0.12} & 3.21 & 0.69 & \textbf{0.83} & 270 & 234 & 37.4 &  784 & 182 & 237\\
 & BiRRT & 32.6 & 26.9 & 0.06 & 125 & 26.6 & 39.7 & 223 & 190 & 0.37 & 746 & 168 & 254\\
 & BiPDGRRT\textsuperscript{2} & 9.16 & 3.76 & 0.25 & 43.6 & 11.1 & 15.8 & \textbf{139} & \textbf{103} & 18.3 & \textbf{438} & \textbf{110} & \textbf{190}\\
 & LightningRRT & 23.8 & 16.0 & 0.07 & 120 & 27.4 & 42.8 & 182 & 136 & 15.7 & 673 & 139 & 205 \\
 \hline
 \multirow{7}{*}{Shelves} & PRM\textsuperscript{1} & 61.9 & 1.28 & 0.90 & 1200 & 194 & 348 & 3001 & 337 & 277 & 28557 & 5670 & 14486\\
 & BiRSG & 1.28 & 1.08 & 0.18 & 5.07 & 0.97 & 1.08 & 16.2 & 14.0 & 1.97 & 47.9 & 9.77 &  11.9\\
 & BiPDGRSG\textsuperscript{2} & \textbf{1.24} &\textbf{ 0.95} & 0.39 & \textbf{4.74} & \textbf{0.91} & \textbf{0.95} & 36.9 &  37.7 &  10.2 &  94.6 &  14.6 &  14.0\\
 & LightningRSG & 2.40 & 2.29 & \textbf{0.11} & 7.67 & 1.81 & 2.20 & 40.3 & 40.8 &  3.53 & 112 & 27.3 &  36.8\\
 & BiRRT\textsuperscript{1} & 114 & 6.73 & 0.28 & 1461 & 284 & 421 & \textbf{10.1} & \textbf{1.11} & \textbf{0.05} & 113 & 23.1 & 46.0\\
 & BiPDGRRT\textsuperscript{1} & 14.7 & 1.48 & 0.58 & 188 & 33.5 & 66.3 & 12.7 & 12.1 & 5.15 & \textbf{21.3} & \textbf{4.09} & \textbf{9.10}\\
 & LightningRRT\textsuperscript{1} & 80.0 & 5.82 & 0.17 & 982 & 198 & 369 & 11.3 & 6.05 & 3.75 & 64.5 & 13.7 & 29.3\\
 \hline
\end{tabular}
\caption{}
\label{tab:1}
\end{table*}

\subsection{Environment distributions}

We conduct experiments in simulation on three distributions of CSpaces. See Figure~\ref{fig:env_examples} for a visual depiction of each.

\subsubsection{RandomPassage} contains a sequence of ($8$) passages for a 2D point robot that are chosen at random to be either at the top or bottom of the environment. This environment distribution was chosen as a canonical difficult problem for planning from scratch methods such as RRT since it contains a sequence of relatively narrow passages at edges of the environment. Intuitively we expect any experience-based guidance method to do rather well since there are $2^8 = 256$ possible environments and most importantly the dynamic obstacles only either exist or don't in a small set of ($16$) locations. 

\subsubsection{Cubicles} contains a set of cubicle-like obstacles for a planar mobile manipulator (6D CSpace consisting of two translation dimensions and four angular dimensions). Each of approximately 9 groups of cubicles can include 1-4 cubicles and can face either up or down. Besides there being many possible environments in the distribution, this CSpace is particularly difficult for its combination of ``mobile'' and ``manipulator'', where there is no clear straight-line distance metric between configurations. Moreover, tasks in this environment are sampled as random inverse-kinematics (IK) configurations where the end-effector is inside a cubicle (notice the base is too large to fit inside), thus placing both starts and goals of tasks inside narrow passages in CSpace. See Section~\ref{sec:ablation} for a discussion of how we compute distance in this space.

\subsubsection{Shelves} is a 3D workspace for a $5$ degree of freedom manipulator containing shelves for the arm to reach into. The distance from the manipulator base, depth of the shelves, number of shelves, and size of the shelves are all sampled uniformly from an interval. Unlike the previous environments, this is a continuous distribution of CSpaces. As above, tasks are sampled as IK configurations whose end effector is halfway inside the shelves. Depending on the depth of the shelves these can be configurations in very narrow regions of CSpace.

\subsection{Algorithms}

We compare our method, PDG, against two PFS algorithms, PRM~\cite{KavrakiSLO96} and BiRRT~\cite{kuffner2000rrt}, and against one path-database planner, Lightning~\cite{berenson2012robot}. For Lightning we do not use DTW to determine which paths go in the database, and instead do parameter tuning as described in the previous section. All our implementations use a highly parallelized GPU-based collision checker based on spherical approximations of the geometries of the robots. This favors fewer large batch calls of collisions over many small batches, and particularly algorithms such as PRM which iteratively validate large sets of nodes and associated edges. 

We test two versions of baseline tree expansion since all three of PDG, Lightning, and BiRRT depend on it. The first version is the classical implementation of RRT~\cite{lavalle1998rapidly} and the second, which we call Random Sample Guidance (RSG), is tailored to our collision checker as it attempts multiple directions for expansion in parallel and selects the closest valid expansion to the random sample. In open space this is essentially just RRT with extra collision checking but in narrow passages this has the effect of exploring perpendicular to obstacles as it \textit{always} expands the search tree even if the random sample is in a direction where there is an obstacle. As such we test the following algorithms: PRM, BiRRT, BiRSG, PDGRRT, PDGRSG, LightningRRT, LightningRSG.

\subsection{Training --- Parameter Tuning via Cross-Validation}
\label{sec:ablation}

We are interested in the Guided Motion Planning problem, and thus assume access to an offline sample from each CSpace distribution. In our case we create a single fixed database of $256$ samples for each of the environment types. We then use this database to do hyper-parameter tuning for \textit{all four} algorithms. Specifically, we hold out a validation set of $64$ environments and do a grid search over parameter settings for randomly chosen tasks in each environment. In the Cubicles environment one of these parameters $\alpha$ is a scaling term for the angular dimensions which affects distance computations through relative importance of the translational and rotational components of the CSpace. 

\subsubsection{PRM} iteratively samples $N$ nodes, connects them to their nearest $K$ edges, and then checks for a path between start and goal. These affect runtime not only through the properties of the algorithm but because our batched collision checking is more efficient with fewer iterations. We try both these parameters on an exponential scale: $N \in [2^9, 2^{10}, 2^{11}, 2^{12}, 2^{13}, 2^{14}, 2^{15}], K \in [2^3, 2^4, 2^5, 2^6]$. The best performing parameters were: RandomPassage $N=2^{9}, K=2^4$, Cubicles $N=2^{10}, K=2^5$, Shelves $N=2^{11}, K=2^5$. 

\subsubsection{BiRRT and BiRSG} iteratively samples in CSpace and extends a max distance of $\tau$ towards the sample from the nearest neighbor on the search tree. RSG does this in batches of extensions of size $b$. Bidirectional search  checks whether the forward and backward trees can be connected if they get within a radius $r_{attach}$ of each other. Finally the CSpace sample is chosen to be the goal with probability $p_{goal}$. We grid search over these parameters: $p_{goal} \in [0.0, 0.01, 0.02, 0.05, 0.1, 0.2], b \in [16, 32, 64, 128]$, and values of $\tau, r_{attach}$ that depend on the CSpace. Note that both Lightning and PDG depend on these algorithms for baseline exploration and thus these same parameters apply to them. The best performing parameters were: RandomPassage $p_{goal} = 0.02, \tau=6.0, b=32, r_{attach}=5.0$, Cubicles $p_{goal} = 0.01, \tau=8.0, b=64, r_{attach}=16.0, \alpha=4.0$, Shelves $p_{goal} = 0.05, \tau=1.0, b=128, r_{attach}=4.0$. 

\subsubsection{PDG} has two main parameters, the set of paths $D_P$ and the radius $\delta$ used to attach to filter them. We try values of $\delta$ that are multiples of the max extension distance $\tau$ above, namely $\delta \in [0.5\tau, \tau, 2\tau, 4\tau]$. We first create a database of $32$ randomly sampled tasks for each environment for a total of $8192$ paths. Next we create $4$ random samples of various sizes as percentages of the original database, $[0.05, 0.1, 0.15, 0.25, 0.5, 0.75]$. We try clustering with DTW thresholds in $[20,40,80,160,200,250,300]$. Finally we also try both interpolated paths (which have many intermediates) and maximally uninterpolated paths (in which no three states are co-linear). The best performing parameters were: RandomPassage $\delta=6.0$ and a database filtered with DTW threshold of $250$, Cubicles $\delta=8.0$ and database filtered with DTW threshold of $40$, Shelves $\delta=2.0$ and database containing $0.25$ percent of the original. 

\subsubsection{Lightning} starts by finding the closest $k$ paths to the start and goal, fully validates these, then runs BiRRT to fix invalid segments. Besides the parameters for BiRRT above we try $k \in [10, 20, 40, 80, 160]$, where the original paper proposed to use $k=10$ but we find higher values worked better with our collision detector. We try the same set of path databases as in PDG above. The best performing parameters were: RandomPassage $k=80$ and a database filtered with DTW threshold of $200$, Cubicles $k=80$ and database containing $0.15$ of the original, Shelves $k=80$ and database filtered with DTW threshold of $7.5$. 

\section{Results}

We report runtime and total number of collision check results for the best parameters for each algorithm on each environment in Table~\ref{tab:1}. Values are computed for $64$ test environments that were not seen during training and for which we sample a single task. We remove the $4$ worse runs for each method. Each method succeeded all $64$ trials in each environment except\textsuperscript{1} PRM and all RRT based methods only succeeded $51$ in Shelves within our $20$ minute time limit. Also in the Shelves environment we found that PDG\textsuperscript{2} performed better when run bidirectionally. In other words, our algorithm (unlike Lightning) is not symmetric with respect to the start and goal, so we can run a bidirectional version which grows two trees from the start and goal respectively and stops if it manages to connect them (just like in BiRRT). The drawback is larger overhead (e.g., initial goal filtering).

\begin{figure}
    \centering
    \includegraphics[width=0.49\linewidth]{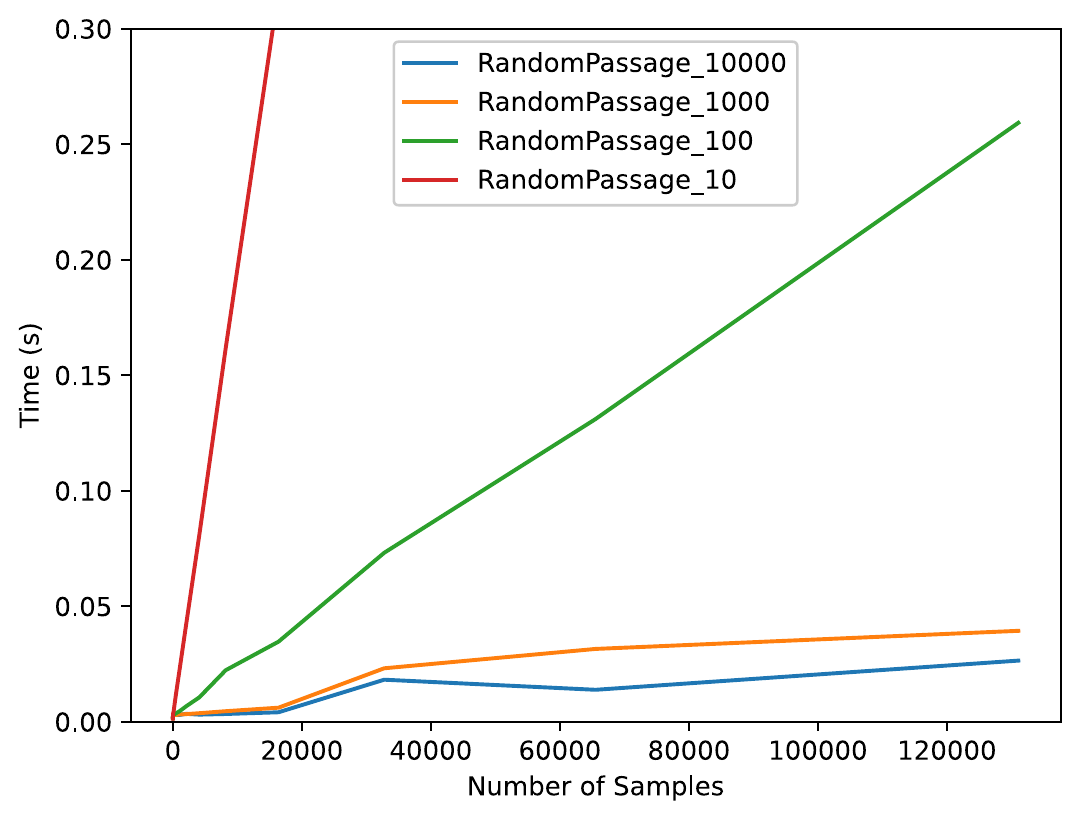}
    \hfill
    \includegraphics[width=0.49\linewidth]{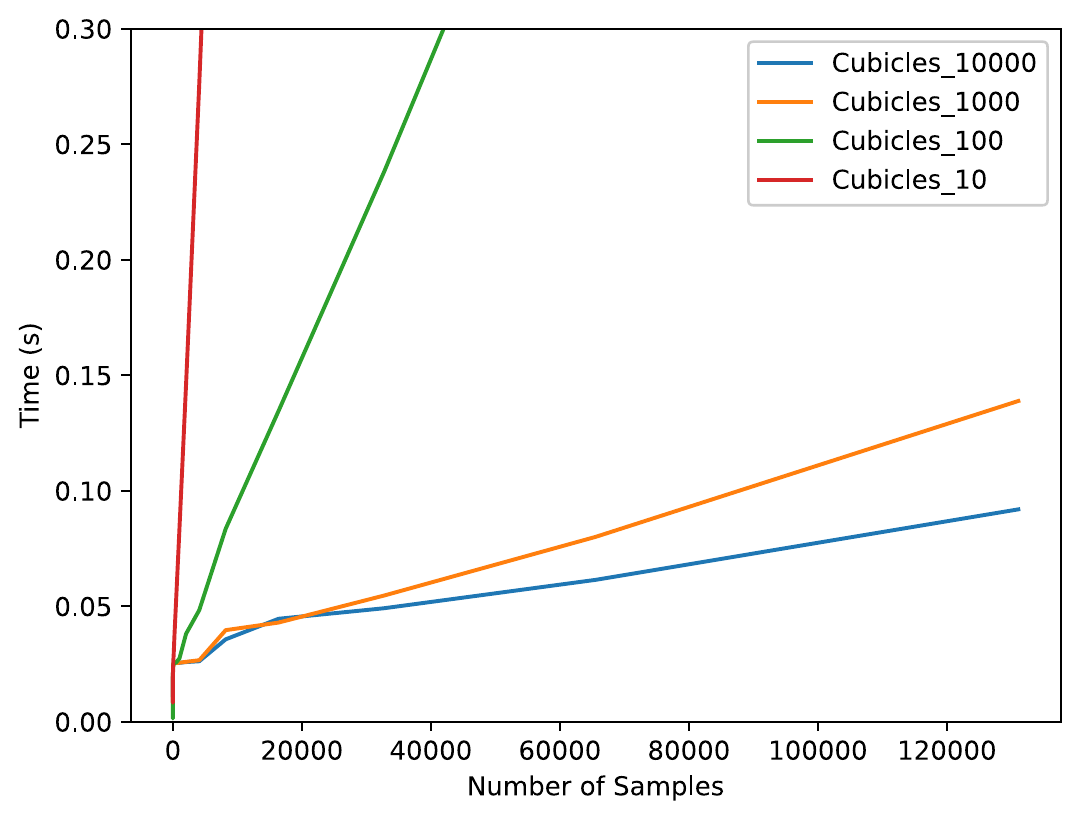}
    \caption{Our CC runtime as a function of number of samples depends greatly on GPU parallelism. At max parallelism (``10000'') the slope for RandomPassage is barely increasing even with a hundred thousand configurations to validate. Overhead in Cubicles, for example moving the data to the GPU, dominates even up to tens of thousands of configurations, meaning that batched calls are significantly more efficient.}
    \label{fig:time}
\end{figure}

Across all three CSpaces PDG is significantly faster than the competitors \textit{when RRT is used for baseline exploration}. We attribute this to the fact that RRT struggles by default in all environments and PDG has relatively less dependence on exploration than the other algorithms. When RSG is used for exploration our method is faster, more consistent, and uses up to an order of magnitude fewer collision checks in both RandomPassage and Cubicles. This is particularly significant because our collision checker is approximate and fully GPU based (including forward kinematics), running on an NVIDIA RTX A5000 with $24$GB of RAM. For example, even with hundreds of thousands of collision checks in batch the slope of the runtime is nearly flat in RandomPassage (see Figure~\ref{fig:time}). This means that with less memory, or no GPU access, or any other reason for a slower collision checker, our runtime results are even more significant. Moreover, both PRM and Lightning do most of their collision checking in large batch calls. Lightning, for which we set $k=80$ as the best performing value from ablations (and $8$ times higher than in~\cite{berenson2012robot}), fully collision checks all queried paths upfront.

In Shelves the best performer (if we again focus on collision checking) is BiRRT (or BiRSG if we focus on methods that did not fail). This matches expectation since tasks in this environment are mainly characterized by their being in long very narrow passages which are difficult to enter or sample (e.g., PRM performance) but relatively easy to escape after which the remaining CSpace is mostly open space (e.g., BiRRT performance). The fact that our runtime remains competitive with BiRSG in this environment indicates that overhead from computing database queries is not too high and provides enough value to make up ground. 

Since all methods terminate upon finding a path we do not focus on path length as an optimization variable, that being said Table~\ref{tab:path_length} shows that the first path found by PDG is much shorter than both BiRRT and Lightning. This matches intuition since PDG uses smoothed paths and Figure~\ref{fig:lightning_bad} shows intuition for why Lightning finds even longer paths than BiRRT since it repairs invalid segments that go in the wrong direction.

\begin{table}[t]
\centering
\begin{tabular}{|c | c | c | c | } 
 \hline 
 & PDG & Lightning & BiDir \\
 \hline
 RandomPassage & \textbf{78.8} & 94.5 & 96.1\\
 \hline
 Cubicles & \textbf{74.5} & 100.8 & 95.7\\
 \hline
 Shelves & \textbf{8.23} & 14.39 & 11.63\\
 \hline
\end{tabular}
\caption{Length of first path found}
\label{tab:path_length}
\end{table}
\section{Limitations}

A limitation of our method is that we operate entirely in CSpace using absolute path configurations. Therefore, while the overall framework of repeatedly querying paths for guidance is general, extending our approach to use more generic path representations (e.g., ego-centric or workspace paths) is left for future work. Another direction for future work is to study whether initially filtering the paths to only those that pass near the goal is always wise, since if we think of global planning as a sequence of local sub-problems then other paths far from the goal can be useful.

A limitation of our results is that our experiments, while covering broad CSpaces, are ``lab grown''. As mentioned above, we believe that the size of our offline training dataset is not larger than is fair in a normal application, and yet recognize that these are nevertheless not real world distributions. We believe the curation of such datasets in robotics is still an open problem, and so focus here on simulation where we can create our own distributions and datasets.

\section{Conclusion}

In this work we argue that within the framework of Guided Motion Planning we should answer questions such as ``which paths should a path database include'' by treating the path database as an input parameter that can be validated during training. Thus we present Path Database Guidance, which uses a database of paths to guide search tree growth along paths that pass near the tree. It has the important property that as the search tree grows the guidance updates through the deletion of invalidated path segments, and when the database fails to provide guidance to any part of the search tree it dynamically swaps to tree expansion according to a baseline exploration method. Finally, PDG performs well in simulation against both an existing standard for planning from experience and against baselines that are well suited for modern massive parallelization (e.g., PRM).

\bibliographystyle{plainnat}
\bibliography{refs}

\end{document}